\documentclass[12pt]{article}
\usepackage{authblk}
\usepackage{amsfonts}
\usepackage{amssymb}
\usepackage{amsmath}
\usepackage{amsthm}
\usepackage{framed}
\usepackage{makecell}

\usepackage{enumitem}
\usepackage{geometry}
\usepackage{url}
\usepackage{orcidlink}
\usepackage{booktabs}

\usepackage{multirow}

\usepackage{mdframed}
\usepackage{mathtools}
\usepackage{algorithm}  
\usepackage{algpseudocode}
\usepackage{array}
\usepackage{amsbsy}
\usepackage{longtable}
\usepackage{caption}

\newcommand{\comb}[1]{\textcolor{black}{#1}}

\newcommand{\C}{\mathbb{C}}  
\newcommand{\Co}{\mathbb{C}}

\newcommand{\anC}{\langle \mathbb{C} \rangle}

\newcommand{\R}{\mathbb{R}}

\newcommand{\x}{x}

\newcommand{\inte}{\mathrm{in}}

\newcommand{\h}{\mathrm{ht}}

\newcommand{\ex}{\mathrm{ext}}

\begin{document} 
\begin{center}
   {\Large\bf 
   
    A Unified Approach to Inferring Chemical Compounds with the Desired Aqueous Solubility}
\end{center} 

\begin{center} 
Muniba Batool$^1$, 
Naveed Ahmed Azam\orcidlink{0000-0002-7941-3419}$^{1, *}$, 
Jianshen Zhu\orcidlink{0000-0002-1287-9572}$^2$,  
Kazuya Haraguchi\orcidlink{0000-0002-2479-3135}$^{2}$, %
Liang Zhao\orcidlink{0000-0003-0869-7896}$^3$, 
 and  
 Tatsuya Akutsu\orcidlink{0000-0001-9763-797X}$^4$ 
\end{center} 
%
%
\begin{center}
{\small 
 $^1$Discrete Mathematics and Computational Intelligence Laboratory, 
 Department of Mathematics, Quaid-i-Azam University, 
 Pakistan\\
 $^2$Discrete Mathematics Laboratory, 
 Department of Applied Mathematics and Physics, Graduate School of Informatics, Kyoto University, 
 Kyoto 606-8501, Japan\\
$^3$Graduate School of Advanced Integrated Studies in Human Survavibility
     (Shishu-Kan),   Kyoto University, Kyoto 606-8306, Japan \\
$^4$Bioinformatics Center,  Institute for Chemical Research, 
  Kyoto University, Uji 611-0011, Japan 

munibabatool@math.qau.edu.pk, 
naveedazam@qau.edu.pk, 
zhujs@amp.i.kyoto-u.ac.jp, 
haraguchi@amp.i.kyoto-u.ac.jp, 
liang@gsais.kyoto-u.ac.jp, 
takutsu@kuicr.kyoto-u.ac.jp \\
$^*$Corresponding author
}

\end{center}

\begin{quote}  
{\bf Abstract}\\ 
Aqueous solubility~(AS) is a key physiochemical property that plays a crucial role in  drug discovery and material design.  
We report a novel unified approach to predict and infer chemical compounds with the desired AS based on simple deterministic graph-theoretic descriptors, multiple linear regression (MLR) and mixed integer linear programming~(MILP). 
Selected descriptors based on a forward stepwise procedure enabled the simplest regression model, MLR, to achieve significantly good prediction accuracy compared to the existing approaches, achieving the accuracy in the range [0.7191, 0.9377] for 29 diverse datasets. 
By simulating these descriptors and learning models as MILPs, we inferred mathematically exact and optimal compounds with the desired AS, 
prescribed structures, and up to 50 non-hydrogen atoms in a reasonable time range 
$[6, 1204]$ seconds. 
These findings indicate a strong correlation between the simple graph-theoretic descriptors and the AS of compounds, potentially leading to a deeper understanding of their AS without relying on widely used complicated chemical descriptors and complex machine learning models that are computationally expensive, and therefore difficult to use for inference. 
An implementation of the proposed approach is available at~\url{ https://github.com/ku-dml/mol-infer/tree/master/AqSol}.

	
	
	\end{quote}
	
\section{Introduction}
The study of quantitative structure-activity/property relationship (QSAR/QSPR) and inverse QSAR/QSPR is crucial in the field of computational chemistry, bio-informatics and material informatics to understand the complex relationships between molecular structures and their properties~\cite{QSAR}.
QSAR/QSPR aims to predict the properties of a given chemical compound, while inverse QSAR/QSPR seeks to infer chemical compounds of desired properties. 

Aqueous solubility (AS) of chemical compounds is a physiochemical property with great significance in various areas such as drug discovery, and material design~\cite{A.S}.
There has been a notable focus on QSPR for the accurate prediction of  AS  through machine learning models such as 
multiple linear regression (MLR), 
logistic regression (LR), 
least absolute shrinkage and selection operator (LASSO), 
partial least square (PLS), and 
random forest (RF). 
We give a brief review of some recent prediction models as follows. 

Palmer et al.~\cite{C9} developed a prediction model based on RF with 
2D and 3D descriptors generated by molecular operating environment. 
Raevsky~et~al.~\cite{C14} represented chemical compounds by the descriptors 
generated from Hybot, Dragon, and VolSurf, and compared the accuracy of prediction models constructed by LR and RF. 
Lowe~et~al.~\cite{C16} utilized the PaDEL-Caret package to generate descriptors,
 and predicted aqueous solubility with RF. 
Lovrić~et~al.~\cite{C13} used LASSO, RF, and light gradient boosting machine (lightGBM). 
For the representation of compounds, they used fingerprints and molecular descriptors generated by Dragon. 
Tayyebi~et~al.~\cite{Tayyebi} used MLR and RF with Mordred package to generate 2D and 3D descriptors.
Wang et al.~\cite{C3} employed LR to construct prediction models for five datasets, where they generated 2D and 3D descriptors by Sybyl and Amber, respectively.
 Meftahi~et~al.~\cite{C2} generated diverse descriptors using Gauusian09, Sybyl, and BioPPSy to predict AS by MLR.
 Cao~et~al.~\cite{Cao} used PLS, back-propagation network (BPN) and support vector regression (SVR) to model  the relationship between molecular descriptors and AS. 
Deng~et~al.~\cite{Deng} used different neural networks such as convolution neural network (CNN),  recurrent neural network (RNN), deep neural network (DNN), and shallow neural network (SNN) with molecular descriptors obtained from Dragon. 
Panapitiya~et~al.~\cite{C10} used RDKit, Mordred, and Pybel for generating descriptors and  employed a graph neural network (GNN) for prediction. 
Hou~et~al.~\cite{C5} proposed a deep learning model named bidirectional long short-term memory with channel and spatial attention network (BCSA) that generates descriptors and construct prediction models. 
Francoeur~et~al.~\cite{C4} presented Sol-TranNet, a molecule attention transformer to predict aqueous solubility. 
They used SMILES representations of molecules and   RDKit to generate descriptors. 
Graph convolution neural networks (GraphConv NN) have been utilized by Conn~et~al.~\cite{C8}.  
 They used descriptors generated by RDKit and Mordred.
Li~et~al.~\cite{Li} developed a model by using cuckoo search algorithm with light gradient boosting machine (CS-LightGBM) where molecular fingerprints are used as molecular representation to express the structure of compounds.
Tang~et~al.~\cite{Tang} introduced self-attention-based message-passing neural network (SAMPN) model. 
They generated specific descriptors by message passing network encoder (MPN), and tested the model on a single dataset. 
A summary of these models is given in Table~\ref{table:Review} 
with 
the number of testing datasets, 
descriptor information, software used to generate descriptors, and the evaluation scores $\mathrm{R}^{2}$, 
where the
minimum and maximum scores are listed if more than one dataset is used in the corresponding model.

\begin{longtable}{*{6}{>{\centering\arraybackslash}p{2.4cm}}}
\caption{A summary of recent models used to predict aqueous solubility.} \label{table:Review} \\
\toprule
S. no & Model & \# datasets & Descriptor information (size) & Software & $\mathrm{R}^2$ Min, Max \\
\midrule
\endfirsthead

\caption{A summary of some recent models used to predict aqueous solubility. (continued)} \\
\toprule
S.no & Model & \# datasets & Descriptor information (size) & Software & $\mathrm{R}^2$ Min, Max \\
\midrule
\endhead

\midrule
\multicolumn{6}{r}{\footnotesize (Continued on next page)} \\
\endfoot

\bottomrule
\endlastfoot

1 & RF~\cite{C9} & 1 & Deterministic 2D, 3D (200) & MOE & 0.89 \\
\midrule
2 & LR, RF~\cite{C14} & 1 & Non-deterministic (21) & Hybot, Dragon, Sybyl, VolSurf & 0.701, 0.736 \\
\midrule
3 & RF~\cite{C16} & 1 & Non-deterministic (16) & PaDEL-Caret package & 0.82 \\
\midrule
4 & LASSO, PLS, RF, LightGBM~\cite{C13} & 1 & Non-deterministic (317) & Dragon & N/A \\
\midrule
5 & MLR, RF~\cite{Tayyebi} & 1 & Deterministic 2D, 3D & Mordred package & 0.80, 0.98 \\
\midrule
6 & MLR~\cite{C3} & 5 & Deterministic 2D, 3D (58) & Sybyl, Amber & 0.4, 0.9 \\
7 & MLR~\cite{C2} & 7 & Deterministic (2, 3, 8)\footnote{Different numbers of descriptors generated by different software.} & Gaussian09 program, Sybyl, BioPPSy & 0.47, 0.87 \\
\midrule
8 & PLS, BPN, SVR~\cite{Cao} & 1 & Deterministic (28) & Dragon & 0.69, 0.735 \\
\midrule
9 & CNN, RNN, DNN, SNN~\cite{Deng} & N/A
& Non-deterministic (N/A) & N/A & N/A\\
\midrule
10 & GNN~\cite{C10} & 1 & Non-deterministic 2D, 3D (839) & Mordred, Pybel, RDKit & 0.76 \\
\midrule
11 & BCSA~\cite{C5} & 5 & Non-deterministic & Within model & 0.83, 0.88 \\
\midrule
12 & Sol-TranNet~\cite{C4} & 5 & Deterministic (25) & RDKit & 0.65, 0.89 \\
\midrule
13 & GraphConv NN~\cite{C8} & 1 & Non-deterministic 2D, 3D (839) & Mordred, Pybel, RDKit & 0.86 \\
\midrule
14 & CS-LightGBM~\cite{Li} & 1 & Non-deterministic & RDKit & 0.8575 \\
\midrule
15 & SAMPN~\cite{Tang} & 1 & Deterministic & MPN & N/A
\end{longtable}

From Table~\ref{table:Review}, we can observe that
most of the models are tested on a single dataset and a few are tested on five or seven datasets, which is very limited size for an in-depth analysis of a prediction model;
some models used non-deterministic 3D and chemical descriptors, making it difficult to use them for the inverse QSAR/QSPR; and
some of the listed models did not achieve good evaluation scores for all the tested datasets, thereby making their applicability to other datasets questionable. 
Furthermore, to the best of our knowledge, no inverse QSAR/QSPR model exists that is specifically designed to infer chemical compounds with the desired AS. 

\comb{
Recently, Azam et al.~\cite{acyclic} proposed a novel framework based on machine learning models and MILP to infer acyclic chemical structures with a desired property value. 
Shi~et~al.~\cite{shi} and Zhu~et~al.~\cite{modeling} extended this framework to infer chemical structures with rings. 
Similarly, Ido~et~al.~\cite{poly} extended the framework for polymers. 
The framework has two phases: prediction phase and inference phase. 
A chemical compound is modeled as a chemical graph. 
Instead of using complicated non-deterministic chemical descriptors that are difficult to compute, and hence challenging for inverse QSPR, simple deterministic graph-theoretic descriptors are developed to construct prediction functions in the prediction phase. 
Other existing inverse QSPR approaches based on heuristics or statistical optimization algorithms do not ensure the exactness and optimality of the inferred chemical compounds, i.e., 
such approaches can infer invalid compounds, and the inferred compounds may not attain the desired property value. 
To avoid such issues in the inference phase of the framework, the descriptors and prediction functions are simulated by MILP formulations  that are  feasible if and only if there exists a desired chemical graph, and thus ensures the exactness and optimality of the inferred chemical graph.
This formulation also allows the users to specify a prescribed structure to be preserved in the inferred graph.} 

\comb{Motivated by the importance of AS in drug discovery and material design, we aim to develop an approach that can address the shortcomings of the existing models. 
For this purpose, we use the framework~\cite{shi, modeling} to:
(i)~accurately predict AS for diverse datasets; and  
(ii)~efficiently infer mathematically exact and optimal chemical compounds with the desired AS.
The efficiency of this framework highly relies on the accuracy of the prediction phase. 
Therefore we modify the framework by introducing 
(a)~a forward stepwise procedure~(FSP) with MLR 
to select significant descriptors which are crucial for achieving high accuracy; and 
(b)~different prediction strategies based on the simplest regression model, MLR, to construct good prediction functions. 
In contrast to the existing approaches, which are tested on a very limited number of datasets, we collected 29 diverse datasets to demonstrate the usefulness of our proposed approach.
The proposed prediction strategies  constructed accurate prediction functions for all 29 datasets and achieved higher accuracy compared to the recent existing approaches for the several datasets.
Furthermore, the approach successfully inferred several chemical graphs with desired AS and prescribed structures in a reasonable time. 
All datasets, source codes and results are available at~\url{ https://github.com/ku-dml/mol-infer/tree/master/AqSol}. }


 \section{\comb{Our} Approach } \label{Our model} 
\begin{figure}[!ht]  \begin{center}
	\includegraphics[width=.95\columnwidth]{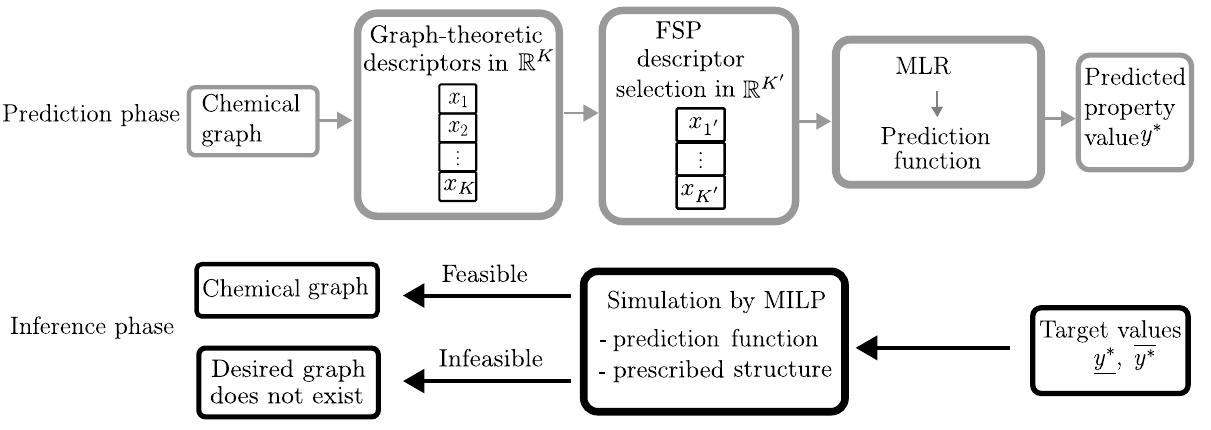}
		\end{center} \caption{An illustration of the  our approach to inferring a chemical graph with the desired AS.} 
		\label{fig:proposed_model}  \end{figure}

\comb{ Our approach is based on the framework~\cite{shi, modeling} to predict and infer chemical graphs with the desired AS. 
 The inference phase of the framework highly depends on the learning performance of the prediction function constructed in the prediction phase. 
 Therefore we modify the framework by introducing an FSP with MLR to select a set of best descriptors,  and different learning strategies based on MLR to construct good prediction functions. 
The details of our approach are discussed in Sections~\ref{phase1} and~\ref{phase2}. 
An illustration of the approach is given in Figure~\ref{fig:proposed_model}. }


\subsection{Prediction Phase} \label{phase1}
\begin{figure}[!ht]  \begin{center}
			\includegraphics[width=.9\columnwidth]{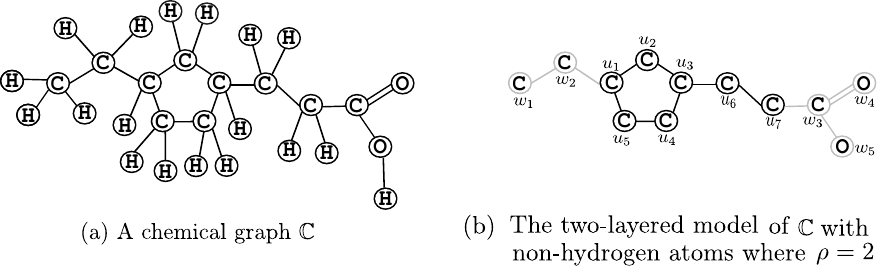}
		\end{center} \caption{(a) Representation of the chemical compound 3-(3-Ethylcyclopentyl) propanoic acid with CID = 20849290 as a chemical graph $\Co$;
		(b) The vertices and edges of the interior and exterior parts of $\Co$ depicted with black and gray colors, respectively, in the two-layered model.
		The sets of interior and exterior vertices are 
		$\{u_1, u_2, \ldots, u_7\}$ and 
		$\{w_1, w_2, \ldots, w_5\}$, respectively. }
		\label{fig:e}  \end{figure}

Modeling: We represent a chemical compound as a chemical graph based on the modeling introduced by 
Zhu~et~al.~\cite{modeling}.
A chemical graph $\Co=(H,\alpha,\beta)$ consists of a simple connected and undirected graph $H$,  a vertex-labeling $\alpha$ that keeps the information of chemical elements, such as {\tt C} (carbon), {\tt O} (oxygen), {\tt H} (hydrogen) and {\tt N} (nitrogen), at each vertex, and an edge-labeling $\beta$ that keeps the  information of single, double, and triple bonds between two adjacent atoms. 
The chemical graph $\Co$ of the compound 3-(3-Ethylcyclopentyl) propanoic acid is illustrated in Figure~\ref{fig:e}(a).
\begin{figure}[!ht]  \begin{center}
			\includegraphics[width=.5\columnwidth]{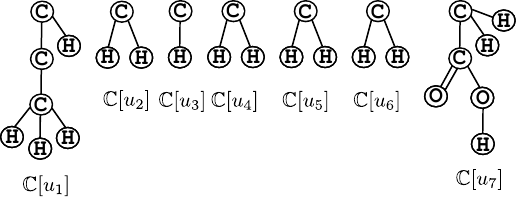}
		\end{center} \caption{The 2-fringe trees $\Co[u_i]$, $i \in [1,7]$ of the example $\Co$ in Figure~\ref{fig:e}(a) rooted at $u_i$.} 
		\label{fig:f}  \end{figure}
\noindent
A chemical graph is divided into {\em interior part} and {\em exterior part} 
based on the 
two-layered model by Shi~et~al.~\cite{shi}. 
For a given parameter $\rho$, the exterior part consists of non-root vertices and edges of rooted tree-like chemical subgraphs called {\em $\rho$-fringe trees} of height at most 
$\rho$. 
Intuitively, the fringe trees resemble terminal functional groups, which play an important role in the properties of the compounds.
The subgraph other than the exterior part of a chemical graph is called the interior part (refer to Appendix~\ref{app:2L} for details). 
The interior and exterior parts of the chemical graph given in Figure~\ref{fig:e}(a) are depicted in Figure~\ref{fig:e}(b), where hydrogen atoms are ignored. 
The 2-fringe trees are illustrated in Figure~\ref{fig:f}. 
		\begin{table}[h!]
			\caption{Descriptors for the chemical graph $\Co $ given in Figure~\ref{fig:e}(a).}\label{table:DC}
			\centering
			\begin{tabular}{@{}l c c c c@{}}
				\toprule
				Descriptor & Descriptor value\\
\midrule
			Number of non-hydrogen atoms in $\Co $ & 12\\
\midrule
                   Rank of $\Co $ & 1\\
\midrule
                   Number of vertices in the interior & 7\\
\midrule
                   Average mass & 56.667 \\
\midrule
                   Number of non-hydrogen vertices of degree 1 & 0 \\
                   Number of non-hydrogen vertices of degree 2 & 5 \\
                   Number of non-hydrogen vertices of degree 3 & 2 \\
                   Number of non-hydrogen vertices of degree 4 & 0 \\
\midrule
                   Number of vertices of degree 1 in the interior& 1\\
                   Number of vertices of degree 2 in the interior & 5 \\
                   Number of vertices of degree 3 in the interior & 1 \\
                   Number of vertices of degree 4 in the interior & 0 \\
\midrule
                   Number of edges of bond multiplicity 2 in the interior & 0 \\
                   Number of edges of bond multiplicity 3 in the interior & 0 \\
\midrule
                   Frequency of chemical elements in the interior: &   \\                  
                   {\tt C} & 7 \\
                   \midrule 
                   Frequency of chemical elements in the exterior: &  \\
{\tt C} & 3\\
{\tt O} & 2 \\
\midrule
                   Frequency of edge-configurations in the interior: &  \\
{\tt C} 2 {\tt C} 2 1 & 2 \\
{\tt C} 2 {\tt C} 3 1 & 5 \\
\midrule
                   Frequency of fringe-configurations in the set of $\rho$-fringe trees: &  \\
{\tt C} 0 {\tt H} 1 & 1\\
{\tt C} 0 {\tt H} 1 {\tt H} 1 & 4\\
{\tt C} 0 {\tt H} 1 {\tt H} 1 {\tt C} 1 {\tt O} 2 {\tt H} 3 & 1\\
{\tt C} 0 {\tt H} 1 {\tt C} 1 H 2 {\tt C} 2 {\tt H} 3 {\tt H} 3 & 1 \\
\midrule
                   Frequency of adjacency-configurations in the set of leaf-edges: & \\
{\tt C} {\tt C} 1 & 1 \\
{\tt O} {\tt C} 1 & 1 \\
{\tt O} {\tt C} 2 & 1 \\

				\bottomrule
			\end{tabular}
		\end{table}

\noindent
Descriptors and their selection: Instead of using some complex chemical descriptors which are hard to compute and use in the inverse QSPR, we use simple and effective graph-theoretic descriptors introduced by Zhu~et~al.~\cite{modeling}. 
For a chemical graph $\Co=(H,\alpha,\beta)$, these descriptors are: 
the number of non-hydrogen atoms in $\Co$;
the rank of $\Co$; 
the number of vertices in the interior;
the average of mass over all atoms in $\Co$; 
the number of vertices of degree $d,  d\in \{1, 2,3, 4\}$ in $\Co$ ignoring the vertices with hydrogen;
the number of vertices of degree $d, d\in  \{1, 2,3, 4\}$ in the interior ignoring the vertices with hydrogen; 
the number of edges with bond multiplicity $m$, $m\in  \{2,3\}$ in the interior;
the frequency of chemical elements in the interior;
the frequency of chemical elements in the exterior;
the frequency of {\em edge-configurations} in the interior which are defined to be the triplets (a$d$, b$d'$,$m$) for each edge $e = uv$ in the interior with 
$\alpha(u)=$ a, $\alpha(v)=$ b, degree of $u$ (resp., $v$) equals to $d$ (resp., $d'$)  and $\beta(e)= m$;
the frequency of {\em fringe-configurations} in the set of ${\rho}$-fringe-trees in  $\C$; and 
the frequency of {\em adjacency-configurations} $({\rm a}, {\rm b}, m)$ in the set of 
leaf-edges
$e = uv$ with either $u$ or $v$ has degree 1 in $\Co$, where  $\alpha(u)=$ a, 
$\alpha(v)=$ b and $\beta(e)= m$.
These descriptors are listed in Table~\ref{table:DC} for an example chemical graph 
$\Co $ given in Figure~\ref{fig:e}(a). 

Selection of significant descriptors plays a key role in constructing good prediction functions. 
We introduce a descriptor selection method  based on the forward stepwise procedure (FSP)~\cite{FSP} and MLR.  
FSP selects significant descriptors iteratively.
That is, it starts with an empty set of selected descriptors, 
at each iteration adds a new descriptor from the set of unselected descriptors that has the optimal MLR evaluation score when combined with the current set of selected descriptors, and terminates the procedure when a desired number of descriptors is selected (refer to Appendix~\ref{FSP} for details). 
We also use LASSO linear regression~(LLR) for descriptor selection in our approach.

\noindent
Prediction strategies: 
We introduce different prediction strategies by using 
FSP for descriptor selection, MLR for prediction, and evaluation methods. 
These evaluation methods mainly depend on leave-one-out validation~(LOOV) and cross validation~(CV)  (refer to Appendix~\ref{app:ML} for details).   
The proposed prediction strategies are listed below:
\begin{itemize}
		\item[-] {FSP-MLR:} FSP is utilized to identify best descriptors, followed by the construction of a prediction function using MLR, and is 
		evaluated by 10 times 5-fold CV.
		\item[-] {FSP-MLR-LOO:} FSP is applied for selecting best descriptors with 5-fold CV for evaluation. 
		Then MLR is employed for prediction, and the performance is evaluated using LOOV.
		\item[-] {FSP-LOO-MLR:} FSP is used for the selection process and MLR is used for the prediction process. Both processes are evaluated by using LOOV. 
\end{itemize}
\noindent
Similarly, we also tried some other prediction strategies based on MLR, LLR and ANN. 
These strategies are listed below:

\begin{itemize}
		\item[-] {MLR:} MLR is applied without selecting descriptors with 10 times 5-fold CV for evaluation.
		\item[-] {MLR-LOO:} MLR is applied without selecting descriptors utilizing LOOV.
		\item[-] {LLR-ANN:} LASSO is used to  identify best descriptors, followed by the construction of a prediction function using ANN. 
		This strategy is evaluated by 10 times 5-fold CV. 
		For more details, we refer to~\cite{modeling}.
		\item[-] {LLR-ANN-LOO:} LASSO is utilized to identify best descriptors followed by the construction of a prediction function using ANN which is evaluated by LOOV.
		This strategy is basically a modification of LLR-ANN~\cite{modeling}. 
		\item[-] {LLR-LLR:} LASSO is utilized for selection of best descriptors and construction of a prediction function. 
		The performance is evaluated by 10 times 5-fold CV. 
		For more details, we refer to~\cite{modeling}.
\end{itemize}	
For all these strategies, we use the graph-theoretic descriptors.

\subsection{Inference Phase} \label{phase2}
\begin{figure}[!h]  \begin{center}
			\includegraphics[width=.90\columnwidth]{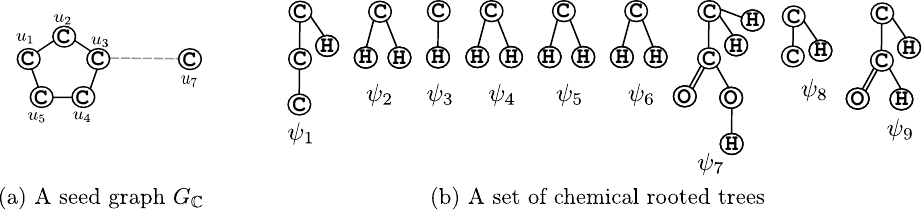}
			\end{center} \caption{
			(a) An illustration of seed graph $G_\Co$ for chemical graph given in Figure~\ref{fig:e}(a) with a typical edge depicted by a dashed line; 
			(b) A set $\mathcal{F} = \{\psi_1, \psi_2, \ldots, \psi_9\}$ of chemical rooted trees, where 
			hydrogen atoms with non-root vertex are omitted.}			
			\label{fig:seed_graph}   
\end{figure}

Several inverse QSPR models are available in the literature. 
However, most of these models heavily rely on heuristic algorithms or statistical optimization techniques, which often result in the inference of invalid compounds or compounds that do not attain the desired property value, and hence can be quite computationally expensive.
In order to avoid such situations, 
we simulate the computation process of a prediction function by an MILP formulation due to Zhu~et~al.~\cite{modeling} to infer chemical graphs with the desired AS. 
A key advantage of this formulation is that it is feasible if and only if a desired chemical graph exists, implying that the inferred graphs will always be valid and achieve the desired AS. Furthermore, this formulation allows users to specify an abstract structure
that is preserved in the inferred graph  by using a
{\em topological specification}. 
A topological specification is described as a set of following rules:
\begin{enumerate}[nosep]
\item[(i)]
a seed graph $G_\Co$ that represents an  abstract form of  a target chemical 
graph $\Co$;
\item[(ii)] 
 a set $\mathcal{F}$ of  chemical rooted trees that are selected for a tree $\C[u]$ with root at each vertex $u$ in the exterior;
and 
\item[(iii)]
lower and upper bounds for the number of components such as vertices in the interior and double/triple bonds within a target chemical graph $\C$.
\end{enumerate} 
For a given seed graph $G_{\Co}$, the formulation constructs the interior of a target chemical graph $\Co$ by replacing typical edges with paths, and exterior by attaching fringe trees. 
Figures~\ref{fig:seed_graph}(a) and~(b) illustrate an example of a seed graph $G_\Co$ with a typical edge, and a set $\mathcal{F}$ of chemical rooted trees, respectively. 
The chemical graph given in Figure~\ref{fig:e}(a) can be obtained from the seed graph $G_\Co$ by replacing the typical edge with a path of length 2, and then attaching the fringe trees from $\mathcal{F}$ accordingly. 
\section{Experimental Results and Discussion}\label{ER}
		We implemented  and tested the proposed approach on a PC with Processor: Core i3 (2.6 GHz at the maximum) and Memory:~6~GB~RAM. 
		\comb{In contract to the existing approaches that are tested on a very limited number of  datasets, we collected 29 diverse datasets to demonstrate the usefulness of our approach. }
		\begin{table}[ht!]\caption{Summary of datasets.} 
			\begin{center}
				\begin{tabular}{@{} c c c c  c  c c c   @{}}\hline
					$\mathcal{C}$ & $ |\mathcal{C}| $ & $ \underline{y},~\overline{y} $ & $| D|$ \\ \hline 
					Protac & 21 & $-6.64,\,-3.18$ & 83\\
					Wassvik & 26 & $-8.49,\,-2.48$ & 94\\
					Alex Manfred & 72 & $-0.833,\,0.65$ & 163\\
					Goodman & 87 & $-6.74,\,-1.06$ & 130 \\
					D5 & 91 &$-5.88,\,0.58$ & 118 \\
					Duffy & 98 & $-10.32,\,-2.48$ & 139\\
					Boobier & 99 & $-8.8,\,1.7$ & 133 \\
					Dearden & 118 & $-6.24,\,-0.57$ & 142 \\
					Ran & 129 &  $-10.8,\,2.06$ & 157 \\
					Llinas & 132 & $-8.75,\,-1.18$ & 167 \\
					Bergstrom & 163 & $-7.59,\,0.55$ & 154\\
					Grigorev & 362 & $-7.85,\,0.38$ & 173\\
					Jain & 456 & $-12.95,\,1.58$ & 223 \\
					Lovric & 805 & $-8.75,\,1.149$ & 323 \\
					Huuskonen & 827 &$ -11.62,\,1.58$ & 310 \\
					David & 826 &$ -10.41,\,1.58$ & 263 \\
					Water set wide & 845 & $-12.79,\,1.58$ & 320\\
					Daniel & 915 & $-10.43,\,6.4$ & 372 \\
					Esol & 1054 & $-11.6,\,1.58$ & 338 \\
					Aqua & 1238 & $-11.62,\,1.58$ & 364 \\
					Tang & 1221 & $-1162,\,1.58$ & 364 \\
					Wang & 1414 & $-9.33,\,1.58$ & 405 \\
					Phys & 1812 & $-12.06,\,1.58$ & 469 \\
					Training set & 5315 & $-13.17,\,2.89$ & 675 \\
					Ochem & 6006 & $-12.1,\,1.58$ & 668 \\
					Cui & 6678 & $-18.21,\,1.7$ & 766 \\
					Aqsol & 8230 & $-13.17,\,2.13$ & 965 \\
					Charles N. Lowe &9150 & $-13.17,\,2.41$ & 835 \\
					Ademola & 10343 & $-13.17,\,2.14$ & 949 \\
					\hline
			\end{tabular}\end{center}\label{table: data}
			 \caption*{$\mathcal{C}$: the dataset; 
			 $|\mathcal{C}|$: the size of $\mathcal{C}$ after the preprocessing;
			$ \underline{y},~\overline{y} $: the lower and upper bounds of 
			AS in each dataset; and 
			$|D|$: the total number of descriptors
			}
		\end{table}
		%
	\smallskip\\
	\noindent
{Datasets: }  
		The 29 diverse datasets are:  
		Protac, Alex Manfred, Ran Yalkowsky, Llinas, Water set wide~\cite{C1},
		Wassvik, Duffy, Dearden, Huuskonen~\cite{C2},
		D5, Jain, Goodman, Wang~\cite{C3},
		Boobier, Aqsol, ESOL~\cite{C4},
		Bergstrom~\cite{C15},
		Grigorev~\cite{C14},
		Lovric~\cite{C13},
		David~\cite{C9},
		Daniel~\cite{C12},
		Tang~\cite{C10},
		Phys, Ochem, Aqua ~\cite{C6},
		Training set~\cite{C8},
		Cui~\cite{C5},
		Charles N. Lowe~\cite{C16}, and 
		Ademola~\cite{C11}.
\smallskip\\
{ Preprocessing: } 		
As a preprocessing, some chemical compounds that do not satisfy one of the following conditions are removed: 
the graph is connected, 
the number of carbon atoms is at least four, and 
the number of non-hydrogen neighbors of each atom is at most 4.
The compounds that are not available in PubChem database~\cite{pub24} are also removed.	
A summary of the datasets is given in Table~\ref{table: data}.  	
These datasets have size in the range $[21, 10343]$, 
AS values in the range $[-13.17, 2.14]$, and 
the number of graph-theoretic descriptors in the range $[83, 965]$.

\subsection{Results on Prediction Phase}\label{ER1}
		Prediction functions are constructed for the 29 datasets based on the prediction strategies by using Python 3.11.3 and Scikit-learn version 1.2.2. 
Based on preliminary experiments, the strategies with LOOV are used for relatively small datasets, with a size of at most 150.	
For such 11 datasets, the $\mathrm{R}^2$ scores due to MLR-LOO, LLR-ANN-LOO, FSP-MLR-LOO, and FSP-LOO-MLR strageties are listed in Table~\ref{FSP-MLR-LOO}. 
For the remaining 18 datasets, 	the results of the prediction strategies MLR, LLR-ANN, LLR-LLR, and FSP-MLR are listed in Table~\ref{MLR}. 
%
\begin{table}[h!]\caption{$\mathrm{R}^2$ scores for small datasets due to MLR-LOO, LLR-ANN-LOO, FSP-MLR-LOO and FSP-LOO-MLR.} 
\setlength\tabcolsep{2pt} 
		\begin{center}
			\begin{tabular}{@{} c c    c  c c c   c   c c c c  @{}}\hline
				$\mathcal{C}$& $|D^*_{\rm L}|$ & $|D^*_{\rm F}|$ & $|D^*_{\rm FL}|$ & $\mathrm{R}^2_{\text{ MLR-LOO}}$ &
	\makecell{$\mathrm{R}^2_{\text{ LLR-ANN-LOO}}$ \\ 
	~\cite{modeling}} &
$\mathrm{R}^2_{\text{ FSP-MLR-LOO}}$ & $\mathrm{R}^2_{\text{ FSP-LOO-MLR}}$ &  $\mathrm{R}^2 $
				\\ \hline
				Protac & 10 & 6 & 10 & $-235.96$ &$^*\mathbf{0.8769}$ & 0.6728 & $^*\mathbf{0.8769}$ & $-0.18$~\cite{C1} \\
				Wassvik & 10 & 2 & 10 & $-0.0125$  & 0.6780 & 0.6788 & $\mathbf{0.8624}$ & 0.95~\cite{C2} \\
				D5 &10& 20 & 25 &$-4 \mathrm{E}10$ &0.7389 & 0.2043 & $^*\mathbf{0.8455}$  & 0.627~\cite{C3} \\
				Alex Manfred &13 & 10 & 30 & 0.044& 0.6573 & 0.647 & $^*\mathbf{0.7593}$ & 0.36~\cite{C1} \\
				Goodman & 18 & 20 & 20 & $-8\mathrm{E}10$ & 0.5147 & $^*\mathbf{0.7830}$ & 0.6363 & 0.527~\cite{C3} \\
				Duffy & 47 & 12 & 45 & $-2\mathrm{E}10 $& 0.7863 & $-0.2177$ & $\mathbf{0.9266}$ & 0.94~\cite{C2} \\
				Boobier & 29 & 12 & 50 & $-8.3\mathrm{E}10$ & 0.6620 & $-0.2140$ & $^*\mathbf{0.8201}$  & 0.773~\cite{C4} \\
				Dearden & 43 & 10 & 55 & $-6.6\mathrm{E}9$ & 0.6987 & $-0.0601$& $\mathbf{0.7191}$ & 0.87~\cite{C2} \\
				Ran & 39 & 22 & 50 &$ -8\mathrm{E}10$ & 0.6719 & $^*\mathbf{0.8931}$ & 0.8041 & 0.82~\cite{C1} \\
				Llinas & 25 & 10 & 35 &$ -2.3\mathrm{E}10 $& 0.5175 &$^*\mathbf{0.7853}$ & 0.6690 & 0.46~\cite{C1} \\
				Bergstrom & 42 & 14& 40 & $-7.6\mathrm{E}10$ & 0.7251 & $-0.0267$ & $^*\mathbf{0.8138}$ & 0.80~\cite{C14} \\
				\hline
		\end{tabular}\end{center}\label{FSP-MLR-LOO}
		\caption*{
	$|D^*_{\rm L}| $: the number of descriptors selected in  LLR-ANN-LOO; 
	$|D^*_{\rm F}| $: the number of descriptors selected in  FSP-MLR-LOO; 
	$|D^*_{\rm FL}| $: the number of descriptors selected in  FSP-LOO-MLR;
$\mathrm{R}^2_{\text{ MLR-LOO}}$: the $\mathrm{R}^2$ score of test data  due to MLR-LOO; 
$\mathrm{R}^2_{\text{ LLR-ANN-LOO}}$: the $\mathrm{R}^2$ score of test data  due to LLR-ANN-LOO; 
$\mathrm{R}^2_{\text{ FSP-MLR-LOO}}$: the $\mathrm{R}^2$ score of test data  due to FSP-MLR-LOO; 
$\mathrm{R}^2_{\text{ FSP-LOO-MLR}}$: the $\mathrm{R}^2$ score of test data  due to FSP-LOO-MLR; 
$\mathrm{R}^2 $: the $\mathrm{R}^2 $ score of the existing model; 
N/A: results not available; 
bold score indicates the best score among our prediction strategies; and 
$^*$ indicates that our best score is better than the scores achieved by the existing models.
		}
	\end{table}	
\begin{table}[h!]\caption{$\mathrm{R}^2$ scores for larger datasets due to MLR, LLR-ANN, LLR-LLR, and FSP-MLR. } 
\setlength\tabcolsep{3.9pt}
		\begin{center}
			\begin{tabular}{@{} c c    c  c c c   c   c c c c  @{}}\hline
				$\mathcal{C}$ & $|D^*|$ & 
				$|D^*_{\text{ FSP-MLR}}|$ & 
				$\mathrm{R}^2_{\rm MLR}$ & 
				\makecell{$\mathrm{R}^2_{\text{ LLR-ANN}}$ \\ ~\cite{modeling}} & 
				\makecell{$\mathrm{R}^2_{\text{ LLR-LLR}}$ \\
 ~\cite{modeling}} 
 & 
				$\mathrm{R}^2_{\text{ FSP-MLR}}$
				& $\mathrm{R}^2$  \\ \hline
				Grigorev &81& 42& $-2.34\mathrm{E}23$ & 0.6685 & 0.6721 & $\mathbf{0.7612}$ & N/A \\
				Jain & 97& 45 &$-1.76\mathrm{E}22$ & 0.9086 & 0.9312 & $\mathbf{0.9377}$ & 0.943~\cite{C3} \\
				Lovric & 66 & 35  &$-6.815\mathrm{E}22$ & 0.7079 & 0.7144 & $\mathbf{0.7294}$ & N/A  \\
				Huuskonen & 141 & 50 & $-1.64\mathrm{E}22$ & 0.8167 & 0.8259 & $\mathbf{0.8371}$ & 0.84~\cite{C2} \\
				David &105 & 43 & $-7.21\mathrm{E}21$ & 0.8410 & $\mathbf{0.8521}$ & 0.8482 & 0.896~\cite{C9} \\
				Water set wide& 81& 51 & $-2.10\mathrm{E}23$ & $^*\mathbf{0.8195}$ & 0.7941 & 0.7975 & 0.77~\cite{C1} \\
				Daniel& 149 & 40 & $-7.899\mathrm{E}24$ & $\mathbf{0.8348}$ &  0.8114 & 0.8264 & 0.935~\cite{C12} \\
				Esol & 222& 60 &$-7.14\mathrm{E}21$ & $\mathbf{0.8659}$ & 0.8147 & 0.8171 & 0.911~\cite{C4} \\
				Aqua & 138& 45 &$-4.75\mathrm{E}23$ & $\mathbf{0.8465}$ & 0.8270 & 0.8399 & N/A \\
				Tang & 154& 60 & $-2.47\mathrm{E}23$  & $\mathbf{0.8487}$ & 0.8211 & 0.8307 & 0.779 \\
				Wang & 145 & 47 & $-5.76\mathrm{E}23$ & $\mathbf{0.8441}$ & 0.7485 & 0.7578 & 0.881~\cite{C3} \\
				Phys & 130  & 60 &$-3.64\mathrm{E}23$ & $\mathbf{0.8867}$ & 0.8287 & 0.8382 & N/A \\
				Training set & 372 & 120 & $-1.24\mathrm{E}23$  & $\mathbf{0.8369}$ & 0.7752 & 0.7776 & 0.86~\cite{C8} \\
				Ochem & 469&110 &$-4.98\mathrm{E}22$ & $\mathbf{0.9313}$ & 0.8405 & 0.8608 & N/A \\
				Cui & 114 & 80& $-9.05\mathrm{E}23$ & 0.7803 & 0.7619 & $\mathbf{0.7806}$ & 0.8813~\cite{C5} \\
				Aqsol& 285 & 113 & $-5.44\mathrm{E}24$ & $\mathbf{0.8184} $ & 0.7198 & 0.7185 & N/A \\
				Charles N. Lowe & 565& 250 &$-1.54\mathrm{E}21$ & $\mathbf{0.8849}$ & 0.7476 &0.7957 & 0.97~\cite{C16} \\
				Ademola & 97 & 180&  $-1.07\mathrm{E}24$  & $\mathbf{0.8675} $& 0.7498 & 0.7830 & N/A \\
				\hline
		\end{tabular}\end{center}\label{MLR}
		\caption*{
		$|D^*|$: the number of descriptors selected in the strategies LLR-ANN and LLR-LLR; 
		$|D^*_{\text{ FSP-MLR}}|$: the number of descriptors selected in FSP-MLR; 
		$\mathrm{R}^2_{\rm MLR}$: the median of $\mathrm{R}^2$ score of test data due to MLR; 
 $\mathrm{R}^2_{\text{ LLR-ANN}}$: the median of $\mathrm{R}^2$ score of test data due to LLR-ANN; 
 $\mathrm{R}^2_{\text{ LLR-LLR}}$: the median of $\mathrm{R}^2$ score of test data  due to  LLR-LLR; 
 $\mathrm{R}^2_{\text{ FSP-MLR}}$: the median of $\mathrm{R}^2$ score of test data due to FSP-MLR; 
 $\mathrm{R}^2 $: $\mathrm{R}^2 $ score of the existing model;
 N/A: results not available; 
 $a {\rm E} b$ represents $ a\!\times\! 10^{b}$;  
bold score indicates the best score among our prediction strategies; and 
$^*$  indicates that our best score is better than the scores achieved by the existing models.
		}
	\end{table}

From Tables~\ref{FSP-MLR-LOO} and~\ref{MLR}, the performance of MLR alone is poor. 
However our FSP-MLR-based strategies  in which  descriptors are selected by FSP and then MLR is applied greatly improved the results. 
Specifically, the best R$^2$ score among the FSP-MLR-based strategies for each of the 29 datasets is at least 0.7198, which falls within the acceptable range, confirming the effectiveness of the proposed strategies.
Similarly, our strategies significantly outperform (resp., yield comparable results to)
the strategies by Zhu et al.~\cite{modeling} for relatively small (resp., large) datasets. 
Notably, FSP-MLR-based strategies achieved the best scores on 16 datasets.
Additionally, our strategies outperform existing results for nine datasets, particularly improving scores for the datasets such as Protac, D5, Alex Manfred, Goodman, and Llinas from $-0.18$, 0.625, 0.36, 0.527, and 0.46 to 0.8769, 0.8455, 0.7593, 0.7830, and 0.7853, respectively. For the remaining 13 datasets with available scores, the results are comparable.
These good evaluation scores are achieved by selecting a small number of descriptors. 
For the small (resp., large) datasets, our model selected descriptors in the range $[6, 39](\%)$ (resp., $[10, 70](\%)$) with an average 21\% (resp., 16\%), which are significantly smaller than those selected by LLR~\cite{modeling} in the range 
 $[10, 70](\%)$ with an average 40\%. 

These experimental results demonstrate that the small numbers of selected graph-theoretic descriptors enabled the simplest regression model MLR to achieve good evaluation scores across the diverse datasets. 
 This indicates a strong correlation between graph-theoretic descriptors and the AS of chemical compounds, paving the way to understanding AS without relying on widely used 3D and chemical descriptors and complex machine learning models, which can be computationally expensive.


\subsection{Results on Inference Phase}\label{ER2}
We selected the datasets Jain and Duffy (resp., Wang and Phys) for which 
FSP-MLR (resp., LLR-ANN) constructed prediction functions with relatively higher evaluation scores. 
For an in-depth analysis, we prepare seven different instances namely
		$I_{\mathrm{a}}$, $I_{\mathrm{b}}^i, i\in\{1, 2, 3, 4\}$, $I_{\mathrm{c}}$
		and $I_{\mathrm{d}}$ with carefully crafted different seed graphs developed by Zhu et~al.~\cite{modeling}. 
		The seed graph of instance $I_{\mathrm{a}}$ is designed to infer any prescribed structures, whereas the seed graphs of instances 
		$I_{\mathrm{b}}^i, i\in\{1, 2, 3, 4\}$	are designed to infer chemical graphs of rank 1 or 2. 
		The seed graphs of instances $I_{\mathrm{c}}$
		and $I_{\mathrm{d}}$ are designed  by merging the structural information of two chemical compounds  obtained from PubChem database~\cite{pub24} to infer a chemical graph that somehow preserves the structure of the two chemical compounds. 
		These instances also heavily depend on other specifications such as the set $\mathcal{F}$ of chemical rooted trees, 
		lower and upper limits for the frequency of chemical symbols, 
		edge configurations and adjacency configurations. 
		We fixed these specifications according to each of the four selected datasets Duffy, Jain, Wang, and Phys. 
		MILP formulations are solved by using {\tt  CPLEX} version 22.1.1. 
		Tables~\ref{table:phase_2_1} and~\ref{table:phase_2_2}	
		(resp.,  Tables~\ref{table:phase_2_3} and~\ref{table:phase_2_4}) show the experimental results of the inference phase for the datasets Jain and Duffy (resp., Wang and Phys). 
	\begin{table}[h!]\caption{Results of the inference phase for the dataset Jain.}
			\begin{center}
				\begin{tabular}{@{}  c  c   c  c c c c c c  c c c     @{}}\hline                
					inst. & $n_{\rm LB}$ &$ \underline{y^*},~\overline{y^*} $ & $\#$v~  &  $\#$c~~   & {\small I-time}\!\! & $ n $ &   $\eta(f(\Co^\dagger))$ &   
					\\ \hline
					$I_{\mathrm{a}}$  & 30 & $-18.75, -18.7$ & 10535 & 9034 & 30.787 & 49 &  $-18.702$\\%
					$I_{\mathrm{b}}^1$ & 35 & $-12.5, -12.45$ & 10402 & 6680 & 11.333 & 35 &  $-12.47$  \\%
					$I_{\mathrm{b}}^2$ & 45 & $-9.95, -9.9$ & 13123 & 9802 & 58.809 & 48 &   $-9.903$ \\%
					$I_{\mathrm{b}}^3$ & 45 & $-13.95, -13.9$ & 12913 & 9804 & 177.04 & 50 &   $-13.907$   \\%
					$I_{\mathrm{b}}^4$ & 45 & $-3.9, -3.85$ & 12707 & 9810 & 110.082 & 50 &  $-3.854$  \\%
					$I_{\mathrm{c}}$ & 50 & $-9.2, -9.15$ & 6651 & 6980 & 6.583 & 50 &    $-9.158$ \\%
					$I_{\mathrm{d}}$  & 40 & $-9.7, -9.65$ & 5271 & 6479 & 67.799 & 44   & $-9.699$  \\
					\hline
			\end{tabular}\end{center}\label{table:phase_2_1}
	\caption*{
						$ n_{\rm LB} $: 
			lower bound for the number of non-hydrogen atoms of target graph $\Co$; 
			$ \underline{y^*},~\overline{y^*} $:  
			lower and upper limits $ \underline{y^*},~\overline{y^*}\in \R$ 
			on the AS of a target graph $\Co $;
			$\#$v and $\#$c are the number of variables and constraints in the MILP,  respectively;
			I-time: the MILP solution time (sec.);
			$n$: the number of non-hydrogen atoms; and 
			$\eta$: the predicted property value 
			 $\eta(f(\Co^\dagger))$ of the inferred chemical graph $\Co^\dagger$.
			}
		\end{table} 
		\begin{table}[h!]\caption{Results of the inference phase for the dataset Duffy.}  
			\begin{center}
			   \begin{tabular}{@{}  c  c   c  c c c c c c  c c c     @{}}\hline                
					inst. & $n_{\rm LB} $ & $ \underline{y^*},~\overline{y^*} $ & $\#$v~  &  $\#$c~~  & {\small I-time}\!\!  & $ n $ &   $\eta(f(\Co^\dagger))$  
					 \\ \hline
					$I_{\mathrm{a}}$ & 30 &$-11.5, -11.45$ &   10535 & 9035 & 16.646  & 42    & $-11.498$  \\
					$I_{\mathrm{b}}^1$ &35 & $-10.9, -10.85 $& 10717 & 6647 & 12.458 & 35 &   $-10.869$ \\
					$I_{\mathrm{b}}^2$ & 45& $-8.4, -8.35$ & 13536 & 9767 & 65.218 &  50 &   $-8.39$ \\
					$I_{\mathrm{b}}^3$  & 45 &$-3.95, -3.9$ & 13337 & 9773 & 31.167 & 50 &   $-3.936$ \\
					$I_{\mathrm{b}}^4$ &45 & $-14.45, -14.4$ & 13138 & 9778 & 91.762 & 49 &   $-14.432$ \\
					$I_{\mathrm{c}}$ & 50 & $-10.1, -10.05$ & 6651 & 6981 & 6.328 & 50 &  $-10.054$ \\
					$I_{\mathrm{d}}$ & 40 & $-8.05, -8$ & 5271 & 6482 & 18.634 & 44 &  $-8.018$  \\
					\hline
			\end{tabular}\end{center}\label{table:phase_2_2}

		\end{table}
		\begin{table}[h!]\caption{Results of the inference phase for the dataset Wang.}
			\begin{center}
				\begin{tabular}{@{}  c  c   c  c c c c c c  c c c     @{}}\hline                 
					inst. & $n_{\rm LB} $ & $ \underline{y^*},~\overline{y^*} $ & $\#$v~  &  $\#$c~~ & {\small I-time}\!\! & $ n $ &  $\eta(f(\Co^\dagger))$ &   
					 \\ \hline
					$I_{\mathrm{a}}$  & 30 & $-6, -5$ & 11212 & 10504 & 25.871 & 35 &  $-5.911$  \\%
					$I_{\mathrm{b}}^1$ & 35 & $-2.5, -2.4$ & 14530 & 8511 & 770.874 &  35 &   $-2.484$ \\%
					$I_{\mathrm{b}}^2$ & 45 & $-2, -1$ & 18475 & 11625 & 1203.539 & 50 &   $-1.063$ \\%
					$I_{\mathrm{b}}^3$ & 45 & $-2.5, -1.5$ & 21717 & 14946 & 251.548  & 20   & $-1.793$ \\%
					$I_{\mathrm{b}}^4$ &45 & $-3, -2$ & 18218 & 11630 & 417.947 & 50 &  $-2.878$ \\%
					$I_{\mathrm{c}}$ &50 & $-4.5, -3.5$ & 7319 & 8435 & 224.792 & 50 &  $-3.836$ \\%
					$I_{\mathrm{d}}$ & 40 & $-5, -4$ & 5942 & 7940 & 13.319 & 42 &  $-4.475$ \\
					\hline
			\end{tabular}\end{center}\label{table:phase_2_3}
		\end{table} 
		
		\begin{table}[h!]\caption{Results of the inference phase for the dataset Phys.}
			\begin{center}
				\begin{tabular}{@{}  c  c   c  c c c c c c  c c c     @{}}\hline                 
					inst. & $n_{\rm LB} $ & $ \underline{y^*},~\overline{y^*} $ & $\#$v~  &  $\#$c~~ & {\small I-time}\!\!   & $ n $ &  $\eta(f(\Co^\dagger))$ &   
					 \\ \hline
					$I_{\mathrm{a}}$ & 30 & $-7.8, -7.75$ & 10880 & 9749 & 376.331 & 45 & $-7.755$    \\%
					$I_{\mathrm{b}}^1$ & 35 &0.15, 0.2 & 14467 & 7855 & 214.926 & 35 &  0.191  \\%
					$I_{\mathrm{b}}^2$ & 45 & 0.52, 0.57& 18326 & 10970 & 1165.992 & 50 & 0.562 \\%
					$I_{\mathrm{b}}^3$ & 45 &$-3.6, -3.55 $& 18166 & 10971 & 322.07 & 50 & $-3.57$ \\%
					$I_{\mathrm{b}}^4$ & 45 & 0.01, 0.06 & 18000 & 10968 & 276.396  & 45 & 0.029  \\%
					$I_{\mathrm{c}}$ & 50 & $-5.47, -5.42$ & 6987 & 7680 & 10.644 & 50 & $-5.436$ \\%
					$I_{\mathrm{d}}$ & 40 &$-2.25, -2.2$ & 5610 & 7185 & 68.318  & 41 & $-2.235$ \\
					\hline
			\end{tabular}\end{center}\label{table:phase_2_4}
		\end{table} 
		\noindent 
	\begin{figure}[!ht]  \begin{center}				
				\includegraphics[width=.97\columnwidth]{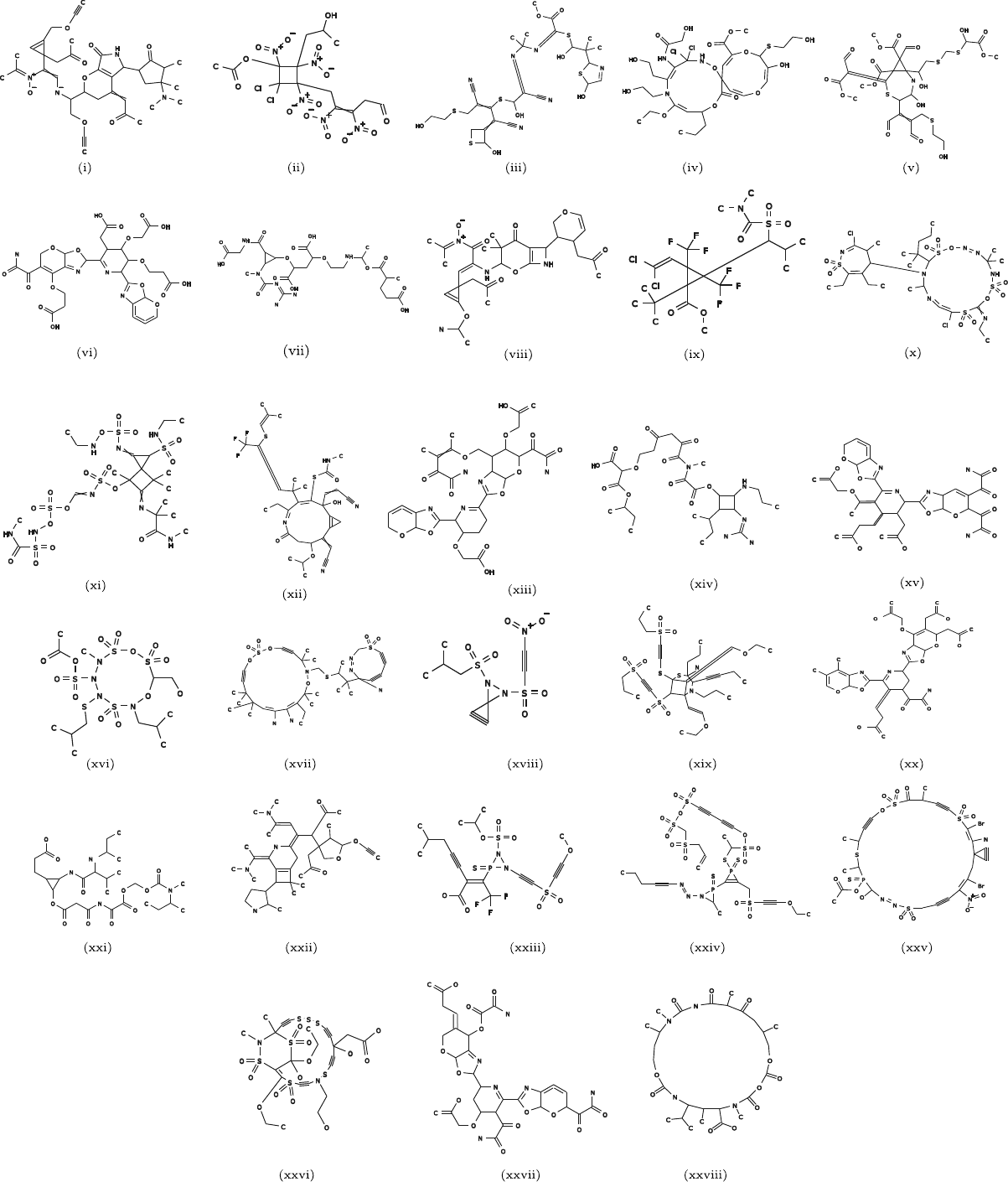}
			\end{center} \caption{
		(i)-(vii), (viii)-(xiv), (xv)-(xxi), and (xxii)-(xxviii) inferred chemical graphs using
			the datasets Jain, Duffy, Wang and Phys, respectively. }
			
			\label{fig:inferred_graphs}   
			\end{figure} 
				
To validate the inferred chemical graph
$\Co^\dagger$,  the AS $\eta(f(\Co^\dagger))$  is also predicted using the corresponding prediction function.  
The experimental results show that even with narrow desired ranges of the AS of the target graphs, the MILP formulations successfully inferred chemical graphs
$\Co^\dagger$ with AS  $\eta(f(\Co^\dagger))$ within the desired ranges while  preserving the prescribed structures, confirming the effectiveness of the MILP formulation.
Additionally, the MILP formulations inferred graphs with relatively larger sizes, with the number of non-hydrogen atoms ranging from 35 to 50, within a reasonable time frame 
[6.328, 1203.539] seconds, demonstrating the efficiency of the inference phase. It is also observed that the instances with a large number of variables and constraints required more time compared to the instances with fewer variables and constraints. 
Furthermore, the MILP solution time when using MLR is significantly shorter than when using ANN. For example, the solution time for the instance 
$I_{\mathrm{b}}^2$ ranges from 11 to 12 seconds with MLR, compared to 1166 to 1204 seconds with ANN. This difference can be due to the lower complexity of the prediction function constructed by MLR compared to that of ANN.
All the inferred chemical graphs are illustrated in Figure~\ref{fig:inferred_graphs}.    
		
%
\section{Conclusion}  \label{CR}
	
A unified approach is proposed to predict and infer chemical compounds with the desired AS. 
Prediction is performed by modeling a chemical compound as a chemical graph with interior and exterior parts which are represented as graph-theoretic descriptors. 
FSP is used to extract significant descriptors followed by MLR to construct prediction functions. 
Graph-theoretic descriptors and prediction functions are simulated by MILPs to infer mathematically exact and optimal chemical graphs with the desired AS and prescribed structure. 

For an in-depth analysis, the proposed FSP-MLR-based prediction strategy was tested on 29 diverse datasets and achieved acceptable evaluation scores for all of the datasets. Our strategies attained significantly higher evaluation scores compared to the recent existing scores, especially improving the scores for the datasets Protac, D5, Alex Manfred, Goodman, and Llinas from $-0.18$, 0.625, 0.36, 0.527, and 0.46 to 0.8769, 0.8455, 0.7593, 0.7830, and 0.7853, respectively. Several chemical graphs with up to 50 non-hydrogen atoms were successfully inferred with the desired AS and prescribed structures for different datasets in a reasonable computation time ranging from 6 to 1204 seconds.
Furthermore, the MILP formulation with MLR-based prediction functions has significantly lower computation time than with ANN-based functions. 
This confirms the effectiveness of our simple approach without relying on complex machine learning models which are quite computationally expensive. 

Experimental results show that the small number of selected graph-theoretic descriptors enabled the simplest regression model, MLR, to achieve high evaluation scores across the diverse datasets, indicating a strong correlation between these descriptors and the AS of chemical compounds. Future work will focus on exploring and investigating the relationships between graph-theoretic descriptors and the AS of compounds that result in a strong correlation, aiming for a better understanding of AS without relying on complicated non-deterministic chemical descriptors.
Furthermore, future work will also focus on efficiently inferring relatively larger chemical graphs, as the computation time of the MILP formulation increases with the number of variables and constraints.

\subsection*{Data Availability}
All datasets, source codes and results are available at \url{ https://github.com/ku-dml/mol-infer/tree/master/AqSol}.

\subsection*{Acknowledgment} 
The publication cost of this article is funded by the Japan Society for the Promotion of Science, Japan,
under grant numbers  22H00532 and 22K19830. 
The second author would like to acknowledge the support provided by JASSO under the JASSO Follow-up Research Fellowship 2024 during his stay at Kyoto University.
\subsection*{Competing interests} The authors declare no competing interest. 
\newpage
\clearpage
 \appendix
\centerline{\bf\LARGE Appendix}

\bigskip
In the following sections, we denote by $V(G)$ and $E(G)$, the vertex set and edge set, respectively, of a graph $G$. 
A path with two end-vertices $u$ and $v$ is denoted by  {\em $u,v$-path}.
The {\em rank}  
of a graph is defined to be 
 the minimum 
 number of edges to be removed so that the graph has no cycles. 
  A {\em rooted} graph is defined to be
 a graph with a  designated vertex, called a {\em root}. 
 The {\em height} of a vertex in a rooted tree is 
 defined to be the size of the longest path from that vertex to a leaf. 
 The {height} 
 of a rooted tree $T$ 
 is defined
to be the maximum height of a vertex in $T$, and is denoted by $\h(T)$.
A chemical compound 
$\Co=(H,\alpha,\beta)$,  
is represented by a simple connected and undirected graph
$H$
 and  
    functions   $\alpha:V(H)\to \Lambda$  and  $\beta: E(H)\to \{1, 2, 3\}$,
    where
 $\Lambda $ is a set of chemical elements such as {\tt C} (carbon), {\tt O} (oxygen), {\tt H} (hydrogen) and {\tt N} (nitrogen). 
 Define the {\em hydrogen-suppressed chemical graph} $\anC$ of a chemical graph 
 $\Co$
to be  the graph obtained from $H$ by
  removing all the vertices whose label is {\tt H}. 
  Figure~\ref{fig:e}(b) illustrates an example of a hydrogen-suppressed chemical graph $\anC$ obtained from the chemical graph $\Co$ of chemical compound 3-(3-Ethylcyclopentyl) propanoic acid given in Figure~\ref{fig:e}(a).

\section{Two-layered model}	\label{app:2L}
%
Let $\Co=(H,\alpha,\beta)$ be a chemical graph and $\rho$ be a positive integer. 
 A two-layered model~\cite{shi} of $\Co$ is defined to be a partition of hydrogen-suppressed chemical graph into interior and exterior as follows. A vertex 
 $v\in V(\anC)$
   (resp., an edge $e\in E(\anC))$ of   $\C$ is called
   an {\em exterior-vertex} (resp.,    {\em exterior-edge}) if $v$ is a non-root vertex of a rooted tree with height at most $\rho$ (resp., $e$ is incident to an  exterior-vertex).
We denote the sets of exterior-vertices and exterior-edges 
by $V^\ex(\C)$ and $E^\ex(\C)$, respectively.
Additionally, we define $V^\inte(\C)=V(\anC)\setminus  V^\ex(\C)$ and 
$E^\inte(\C)=E(\anC)\setminus E^\ex(\C)$, to denote the sets of interior vertices and interior edges, respectively.
 The set  $E^\ex(\C)$ of  exterior-edges forms 
a collection of connected graphs each of which is
considered as a rooted tree $T$ rooted at 
the vertex $v\in V(T)$ with maximum $\h(v)$.
Let $\mathcal{T}^\ex(\anC)$ denote 
the set of these chemical rooted trees in $\anC$. 
The {\em interior} $\C^\inte$ of $\C$ is defined to be the subgraph
 $(V^\inte(\C),E^\inte(\C))$ of $\anC$. 
  For $\rho$ = 2, and the example $\anC$ given in Figure~\ref{fig:e}(b), the interior can be obtained by iteratively removing the set of vertices with degree 1 two times, where 
  $\{w_1, w_2, \ldots, w_5\}$ and 
  $\{u_1, u_2, \ldots, u_7\}$. 
 For each $u\in V^\inte(\C)$,
let $T_u\in \mathcal{T}^\ex(\anC)$ denote the chemical tree rooted at $u$, and we define the $\rho$-fringe tree to be the chemical rooted tree obtained from $T_u$ by putting back the hydrogens. Figure~\ref{fig:f} illustrates the set of 2-fringe trees of the chemical graph $\Co$ in Figure~\ref{fig:e}(a).						

\section{Descriptor Selection} \label{FSP}

Let $\mathcal{C}$ be a dataset of chemical graphs $\Co$, and $a(\Co)\in \R$ 
denote observed value of aqueous solubility of $\Co$.
Let $D$ be a set of descriptors and $f$ represents a feature function that assigns a vector $ f(\Co)=x \in \R^{|D|}$ to a graph $\Co$. The value of descriptor $d\in D$ is denoted by $x(d)$. 

An algorithmic description of the descriptor selection method with FSP and MLR is given in  Algorithm~\ref{alg:FSP}, 
where for a subset 
$D^*\subseteq D$,
		 $\mathrm{R}^2_{\mathrm{MLR}}(\eta,D^*)$ denote the
		  $\mathrm{R}^2$ score of a prediction function 
		  $\eta$ obtained by MLR using descriptor set $D^*$, and for an integer $K\geq 1$, $h_K:2^D\to \R$ is an evaluation function such that $h_K(D^*)=\mathrm{R}^2_{\mathrm{MLR}}(\eta,D^*)$. 
%
%
\begin{algorithm}[htbp]
			\caption{Forward stepwise procedure with MLR}
			\begin{algorithmic}
				\State \textbf{Input:} A set of compounds 
				$\mathcal{C}$, a descriptor set $D$, an integer $K \geq 1$,
				 and a function for evaluation $h_K$ based on MLR.
				\State \textbf{Output:} A descriptor subset $ D^* \subseteq D $.
				\State $ D_{\mathrm{best}} :=  \emptyset $;
				\While{ $|D_{\mathrm{best}}| \neq K$}
				\For{each descriptor $d \in D \setminus  D_{\mathrm{best}}$}
					\State Compute $h_K( D_{\mathrm{best}} \cup \{d\})$
									\EndFor;
					\If{  $ d^* $ maximizes 
					$h_K(D_{\mathrm{best}} \cup \{d\})$ over all $ d \in D \setminus  D_{\mathrm{best}}$ }
					\State  $ D_{\mathrm{best}} :=  D_{\mathrm{best}} \cup \{d^*\}$
				
					\EndIf

				\EndWhile;
\State Return $D_{\mathrm{best}}$ as $D^*$.				
				
			\end{algorithmic}
			\label{alg:FSP}
		\end{algorithm}

\section{Machine learning models and evaluation}\label{app:ML}
We use linear regression and ANN to construct a prediction function which are briefly explained as follow.
\bigskip\\
\noindent 
{Linear regression:} It tries to identify the best linear relationship between the given data points and their observed values. 
More precisely, it finds a hyperplane that minimizes the error between the observed values and predicted values of the given dataset. 
LLR~\cite{LLR} uses an extra regularization term with the error function to  penalize coefficients of the hyperplane. 
Thus LLR selects a hyperplane that minimizes the following LASSO function:
\[\frac{1}{2|\mathcal{C}|}\mathrm{Err}(\eta_{w, b}; \mathcal{C})
+   \lambda( \sum_{j =1}^{|D|}  |w(j)| +  |b| ), \]
where 
$\mathcal{C}$ is the dataset of chemical graphs with $a_i$ and 
$x_i$ to be the observed value and feature vector, resp., 
of $\Co_i \in\mathcal{C}$;
$w, b$ are the coefficients of the constructed hyperplane, 
$\eta_{w, b}$ is the prediction function due to the hyperplane $w, b$;
$\lambda$ is the regularization term, and 
 $\mathrm{Err}(\eta;\mathcal{C}) $  is the error function such that
\[ \mathrm{Err}(\eta_{w, b};\mathcal{C})  \triangleq 
		\sum_{\Co_i\in \mathcal{C}}(a_i - \eta_{w, b}(\x_i))^2.\]
MLR can be considered as a special case of LASSO function when $\lambda = 0$.  \\
\bigskip\\
\noindent 
{Artificial neural network:} 
%
An ANN consists of three kinds of layers: an input layer, hidden layers, and an output layer, with each layer consisting of nodes. 
There are weighted edges between every two consecutive layers and a bias term for each node. 
In this work, we use fully connected feed-forward ANN with rectification linear unit (ReLU) and identity function
as activation functions in the hidden layers and output layer, resp., where 
$ {\rm ReLU}(x) \triangleq {\rm max}(0, x). $
The input layer takes the input data, and the output layer provides the predicted value. 
At each node of the hidden and output layers, the computation of the weighted sum followed by the application of an activation function is performed. 
The ANN algorithm tries to update the weights and biases to minimize the error between observed and predicted values.
\bigskip\\
\noindent 
{Evaluation:} 
Let $K$ denote that number of  descriptors used. 
To evaluate the performance of prediction functions  $\eta: \R^K\to \R$
constructed by our model we define 
		{\em coefficient of determination} $\mathrm{R}^{2}(\eta;\mathcal{C}) $ as
		
		\[ \displaystyle{ \mathrm{R}^{2}(\eta; \mathcal{C} ) \triangleq 
			1- \frac{\mathrm{Err}(\eta;\mathcal{C}) } 
			{\sum_{ \Co_i\in \mathcal{C}  } (a_i-\widetilde{a})^2} 
			\mbox{   for  }
			\widetilde{a}= \frac{1}{|\mathcal{C}|}\sum_{ \Co_i \in \mathcal{C} }a(\Co_i)  }. \]. 
		
We perform 5-fold CV and LOOV as follows: 
		\begin{itemize}
			\item[\textbf{1.}] { 5-fold CV}:  A random partition is made for a set of graphs $\mathcal{C}$ into five subsets  $\mathcal{C}^{(k)}$, $k\in \{1, \dots, 5\}$. For each $k\in\{1, \dots, 5\}$, let $\mathcal{C}_\mathrm{train}:=\mathcal{C} \setminus \mathcal{C}^{(k)}$ and $\mathcal{C}_\mathrm{test}:=\mathcal{C}^{(k)}$. We then construct the prediction function  $\eta^{(k)}: \R^{K}\to \R$ using $\mathcal{C}_\mathrm{train}$ and calculate the score $\mathrm{R}^2(\eta^{(k)},\mathcal{C}_\mathrm{test})$. 
			This process is repeated for $p$ times, and evaluate the model based on the median of $5p$ $\mathrm{R}^2(\eta^{(k)},\mathcal{C}_\mathrm{test})$ scores.
			\item[\textbf{2.}] {LOOV:} 
			For every $i\in \{1, \ldots, |\mathcal{C}|\}$, let $\mathcal{C}_\mathrm{train}:=\mathcal{C} \setminus \{\Co_i\}$, 
			$\mathcal{C}_\mathrm{test}:=\{\Co_i\}$. 
			Then, construct the prediction function  $\eta^{(i)}: \R^K\to \R$ based on $\mathcal{C}_\mathrm{train}$,  and 
			\comb{calculate $\mathrm{R}^2$ score as 
			\[ \displaystyle
						1- \frac{
				\sum_{\Co_i\in \mathcal{C}}(a_i - \eta^i(\x_i))^2 } 
			{\sum_{ \Co_i\in \mathcal{C}  } (a_i-\widetilde{a})^2} 
			 .\].}
		\end{itemize}

\end{document}